\documentclass[sigconf]{acmart}

\usepackage{multirow}
\usepackage{hyperref}
\hypersetup{
    colorlinks=true,
    linkcolor=blue,
    filecolor=blue,      
    urlcolor=blue,
    citecolor=cyan,
}

\AtBeginDocument{%
  \providecommand\BibTeX{{%
    \normalfont B\kern-0.5em{\scshape i\kern-0.25em b}\kern-0.8em\TeX}}}

\acmConference[Conference '17]{Proceedings of the ACM Conference}{July, 2017}{Washington, DC, USA}
\acmBooktitle{Proceedings of the ACM Conference (Conference '17)}
\acmPrice{15.00}

\begin{document}

\title{Recapture as You Want}


\author{Chen Gao}
\affiliation{%
  \institution{IIE, CAS}}
\email{gaochen@iie.ac.cn}

\author{Si Liu}
\affiliation{%
  \institution{Beihang University}}
\email{liusi@buaa.edu.cn}

\author{Ran He}
\affiliation{%
  \institution{Institute of Automation, CAS}}
\email{rhe@nlpr.ia.ac.cn}

\author{Shuicheng Yan}
\affiliation{%
  \institution{Yitu Technology}}
\email{shuicheng.yan@yitu-inc.com}

\author{Bo Li}
\affiliation{%
  \institution{Beihang University}}
\email{boli@buaa.edu.cn}


\begin{abstract}
  With the increasing prevalence and more powerful camera systems of mobile devices, people can conveniently take photos in their daily life, which naturally brings the demand for more intelligent photo post-processing techniques, especially on those portrait photos. 
  In this paper, we present a portrait recapture method enabling users to easily edit their portrait to desired \emph{posture/view}, \emph{body figure} and \emph{clothing style}, which are very challenging to achieve since it requires to simultaneously perform non-rigid deformation of human body, invisible body-parts reasoning and semantic-aware editing. 
  We decompose the editing procedure into semantic-aware geometric and appearance transformation. 
  In geometric transformation, a semantic layout map is generated that meets user demands to represent part-level spatial constraints and further guides the semantic-aware appearance transformation.
  In appearance transformation, we design two novel modules, Semantic-aware Attentive Transfer (SAT) and Layout Graph Reasoning (LGR), to conduct \emph{intra-part transfer} and \emph{inter-part reasoning}, respectively.
  SAT module produces each human part by paying attention to the semantically consistent regions in the source portrait. It effectively addresses the non-rigid deformation issue and well preserves the intrinsic structure/appearance with rich texture details.
  LGR module utilizes body skeleton knowledge to construct a layout graph that connects all relevant part features, where graph reasoning mechanism is used to propagate information among part nodes to mine their relations. In this way, LGR module infers invisible body parts and guarantees global coherence among all the parts.
  Extensive experiments on DeepFashion, Market-1501 and in-the-wild photos demonstrate the effectiveness and superiority of our approach. Video demo is at: \href{https://youtu.be/vTyq9HL6jgw}{\color{magenta}{https://youtu.be/vTyq9HL6jgw}}.
  \end{abstract}



\begin{teaserfigure}
  \includegraphics[width=\textwidth]{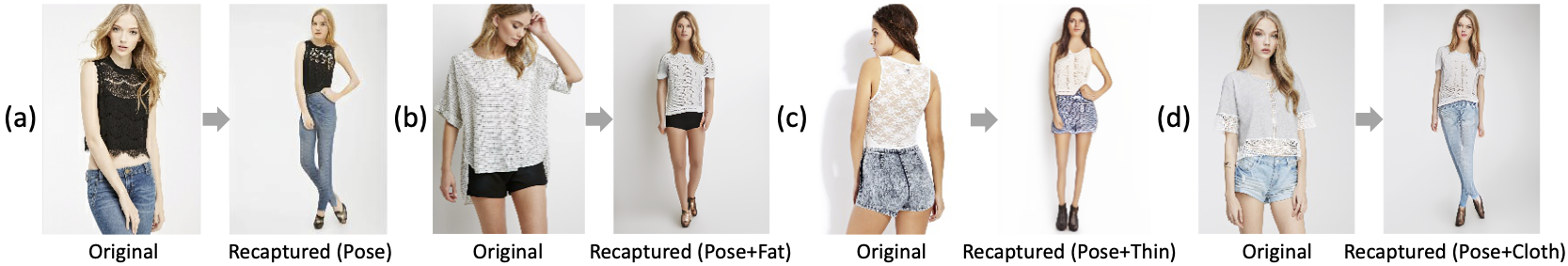}
  \caption{Example recaptured portrait photos with our method. Our method allows users to recapture their portrait photos with desired posture/view, figure and clothing style. (a)$\rightarrow$(d): Recaptured portraits show different posture/view from the original. (b) and (c): Original and recaptured portraits have different figures (fat or thin). (d): Original and recaptured portraits have different clothing. Moreover, when producing recaptured portraits, our method accurately preserves person identity and texture details in terms of both intrinsic structure and appearance. It can properly infer invisible body parts and clothes in original portraits, e.g. the lower body, and meanwhile guarantee global coherency of different regions in recaptured portraits.}
  \Description{...}
  \label{fig:teaser}
\end{teaserfigure}

\maketitle

\section{Introduction}

Nowadays the application of camera systems on mobile devices has been increasingly mature and people can conveniently take photos in their daily life, especially portrait photos. However, the captured photos are usually not so ``perfect'' as people consider they should be. 
For example, a lady may be dissatisfied with their original posture thus wishes she had taken the portrait in another pose or view for showing better figure (Fig.~\ref{fig:teaser} (a)), or  
want to make herself look fatter (Fig.~\ref{fig:teaser} (b)) or thinner (Fig.~\ref{fig:teaser} (c)), 
or change the clothing in the original portrait to another style (Fig.~\ref{fig:teaser} (d)).
People always have endless desire for beauty in their appearance. 
We are thus motivated to build a recapture model that allows the users to easily manipulate their portrait photos according to their specific needs in short time conveniently, rather than artificially editing photos with professional software like PhotoShop. 
In the future, we also expect such a recapture application can work well on mobile devices as a portrait photo customizable editing tool for better user experience.
Therefore, in this work, we introduce an intelligent recapture method that can revise user portrait to desired posture/view, body figure and clothing style simultaneously.

It is very challenging to achieve our goals, mainly due to following three aspects. 
1) \emph{Non-rigid deformation of human body}. To change human pose like Fig.~\ref{fig:teaser}, a model should guarantee that the identity and texture details accord with those of the original person in terms of both intrinsic structure and appearance, which is difficult due to the non-rigid nature of human body. 
Most image editing approaches based on cGANs~\cite{mirza2014conditional} or conditional VAEs~\cite{Kingma2013AutoEncodingVB} are not directly applicable in this setting, 
since simply stacking convolution layers cannot solve spatial misalignment issue caused by non-rigid deformation. Even pose transfer methods~\cite{ma2017pose,si2018multistage,Siarohin2018DeformableGF,zhu2019progressive} can not effectively solve this difficulty. 
For example, Def-GAN~\cite{Siarohin2018DeformableGF} applies affine transformation for each body part to align features, where each part is artificially defined as a rough rectangular region and all the pixels inside roughly share one set of transformation parameters. 
2) \emph{Invisible body-parts reasoning}. For changing the portrait view angle, the model should be able to appropriately infer invisible parts in the original image (e.g., lower body in Fig.~\ref{fig:teaser}) according to the limited visible area, and meanwhile the inferred part should be coherent with the whole human body, such as left and right shoes synthesized to have the same color. Some warp-based methods~\cite{dong2018soft} only consider aligning existing feature and ignore the importance of correctly inferring the invisible parts while guaranteeing the global coherency. 
3) \emph{Semantic-aware editing}.
For precisely changing the body figure and clothing according to use demand, like making legs thinner in Fig.~\ref{fig:teaser} (c) or changing into long trousers in Fig.~\ref{fig:teaser} (d), the model should  be able to perceive the semantic layout of the portrait. The editing process must be semantic-aware and conditioned on this layout. Existing related methods~\cite{si2018multistage,Siarohin2018DeformableGF,zhu2019progressive} only consider body key-points representation and cannot precisely modify specific body parts at semantic-level.

In this work, we propose a GAN-based customizable recapture method to address the above challenges with semantic-aware geometric transformation and appearance transformation.
On one hand, the semantic-aware geometric transformation is proposed to address the non-rigid deformation and semantic-aware editing challenges. It produces a part-level semantic layout map to represent details of desired posture/view and clothing category without appearance information. which the user can interactively edit for obtaining satisfactory body figure. Such a layout map is a high-level intermediate representation and can guide the following appearance transformation with semantic-level spatial constraints.
On the other hand, for appearance transformation, we devise a Semantic-aware Attentive Transfer (SAT) module to further solve the non-rigid deformation challenge and well preserve intrinsic appearance structure of the original person. With SAT, \emph{intra-part transfer} is achieved, which precisely transfers features from source to target region with semantic-level constraints generated by geometric transformation. SAT produces one pixel by exploiting all features from the corresponding region in the original portrait and learning to combine these features via an attention mechanism, which is a fine-grained transformation paradigm capable of tackling non-rigid deformation.
Besides, in appearance transformation we also apply a Layout Graph Reasoning (LGR) module to address the invisible body-parts reasoning challenge. It performs \emph{inter-part reasoning} to leverage hierarchical relations of body parts and the semantic layout to gather relevant features in order to construct a layout graph. Node features propagate on such a graph through GCN~\cite{Kipf2016SemiSupervisedCW} for strengthening the global reasoning capability of our model and enabling more intrinsic relations to be mined among body parts, such as the common similarity of left and right shoes. The hierarchical knowledge and reasoning process provides rich cues that benefit the inference of invisible parts reasonably and help guarantee their global coherency.

To sum up, our main contributions are three-fold. 
1) We present a customizable recapture method with two specially designed modules that complement each other from both intra-part and inter-part perspective, with promising results achieved. 
2) We propose a novel Semantic-aware Attentive Transfer (SAT) module in our network that can effectively learn the transformation between semantically corresponding body parts, which solves the non-rigid deformation of human body and well preserves the intrinsic appearance structure. 
3) We propose a novel Layout Graph Reasoning (LGR) module in our network that can effectively infer the invisible body parts and guarantee the global coherency among different parts via applying global reasoning over hierarchical relation knowledge.

\begin{figure*}
  \begin{center}
  \includegraphics[width=1\linewidth]{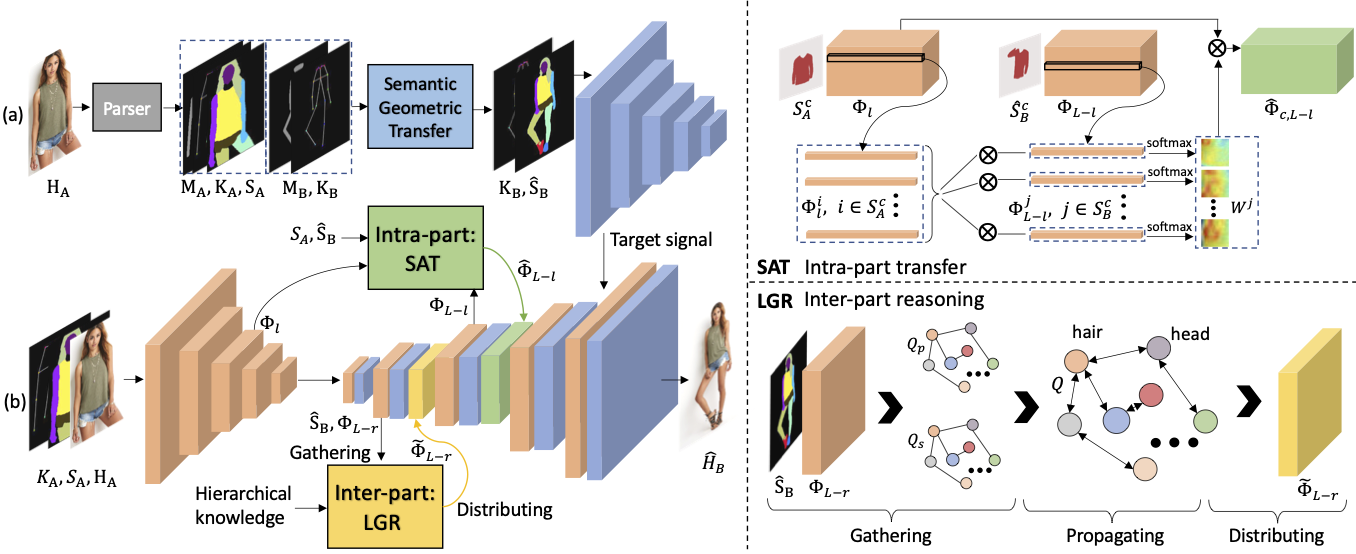}
\end{center}
  \caption{Illustration of our framework, where two proposed modules (SAT $\&$ LGR) are presented in detail on the right.}
  \vspace{-1mm}
  \label{fig:framework}
\end{figure*}

\section{Related Work}
\label{gen_inst}
\subsection{Image Generation} 
Image generation is a fundamental task in computer vision. 
Restricted Boltzmann machines~\cite{hinton2006reducing}, autoencoders~\cite{Kingma2013AutoEncodingVB,sohn2015learning} and flow-based generative models~\cite{dinh2014nice,kingma2018glow} are all efficient approaches. 
In recent years, GANs~\cite{goodfellow2014generative} based methods have also been widely used for synthesizing realistic images and achieved good results in various tasks like noise-to-image~\cite{arjovsky2017wasserstein,mao2017least,karras2017progressive,gao2019adversarialnas}, image-to-image translation~\cite{isola2017image,zhu2017unpaired,wang2018high,choi2018stargan,Jiang2020PSGAN,zhu2019ugan}, 
text-to-image translation~\cite{zhang2017stackgan,xu2018attngan,bodla2018semi} and image inpainting~\cite{pathak2016context,huang2017stacked}. Most image-to-image translation models are based on cGANs~\cite{mirza2014conditional} that can make the translation process more controllable. However, these methods require pixel-wise alignment between input and output images, and hence are not directly applicable to the setting of this work challenged by an inherent difficulty of spatial non-grid deformation.

\subsection{Pose-Guided Person Image Generation}
View synthesis is an important sub-area of image synthesis. Most works on view synthesis focus on simple rigid objects such as furniture and car~\cite{ji2017deep,kulkarni2015deep}. Comparatively, pose-guided person image generation is a more challenging task since it needs to tackle the non-rigid human body. PG\textsuperscript{2}~\cite{ma2017pose} firstly proposes this task, which uses a coarse-to-fine strategy and a common U-Net structure. It gets poor results since it does not consider the spatial misalignment caused by different postures. 
Def-GAN~\cite{Siarohin2018DeformableGF} takes the human body as an articulated object and applies affine transformation to each body part to align features to target pose distribution. However, it defines body part with a rough rectangular region and achieves poor results when generalized to rare human body postures. On the contrary, our method considers the pixel-level spatial mapping between different pose distributions to better preserve the intrinsic appearance structure in the non-rigid deformation setting. 
PATN~\cite{zhu2019progressive} introduces a series of effective pose-attention blocks that replace the residual block in bottleneck layers. It allows the network to generate the human body progressively via each block focusing on specific areas. The difficulty of the transfer process is simplified through such a progressive mechanism. However, it cannot effectively preserve the person identity and texture details in some hard cases.
Previous methods, especially warp-based methods~\cite{dong2018soft}, only consider aligning visible region and ignore the issue of correctly inferring invisible parts while maintaining global coherency. Compared with them, we effectively reason the intrinsic relations among body parts to better address this issue. More importantly, they are not capable of changing the body figure and clothing style, thus cannot directly achieve our goal of portrait recapturing.
\section{Proposed Method}
\subsection{Notations and Overview}
To train our model in a supervised manner, we adopt DeepFashion~\cite{liu2016deepfashion} and Market-1501~\cite{zheng2015scalable} datasets which offer paired data, i.e. a same person in different poses.
$H_A, H_B \in \mathbb{R}^{3 \times H \times W}$ denote the given source (A) human image and target (B) human image, respectively. During training phase, $H_A, H_B$ are randomly sampled from the set of images of the same person in different poses. We adopt the HPE~\cite{cao2017realtime} used by~\cite{Siarohin2018DeformableGF,zhu2019progressive} to estimate $18$ key points of human body and take $K_A, K_B \in \mathbb{R}^{18 \times H \times W}$ as the $18$ channels binary heatmap, where each heatmap is filled with 1 in a radius of 4 pixels around the corresponding keypoints and 0 elsewhere. Morphological operations~\cite{ma2017pose} is utilized to generate the pose mask $M_A, M_B \in {\{0, 1\}}^{H \times W}$. We adopt a human parser~\cite{ruan2019devil} trained on LIP \cite{gong2017look} to extract the parsing maps with twenty semantic categories and represent the map using a one-hot 20-dim vector for each pixel: $S_A, S_B \in {\{0, 1\}}^{20 \times H \times W}$. For Market-1501, we cluster twenty classes into seven classes due to its low image resolution. 

The goal of the generator in our model is to produce a new portrait $\hat{H}_B$ according to a given portrait ${H}_A$ and some requirements such as desired posture/view $K_B$, body figure and clothing style. For training, the generator receives as input the source $\{H_A, S_A, K_A\}$ and the target pose $\{K_B\}$. We decompose the generation procedure to semantic-aware geometric and appearance transformation, as shown in Fig.~\ref{fig:framework} (a) and (b), respectively. On one hand, for geometry translation, a target semantic map $\hat{S}_B$ is generated that represents the target pose $K_B$ by taking the $\{S_A, K_A, K_B\}$ as input. $\hat{S}_B$ contains semantic category and spatial layout information. During inference, $\hat{S}_B$ can be accordingly modified by the users to meet their demands. On the other hand, for appearance translation, the model takes $\{H_A, S_A, K_A, \hat{S}_B, K_B\}$ as input and generates the final image $\hat{H}_B$ with the guidance of semantic-level spatial constraints $\{S_A, \hat{S}_B\}$.
For this part of procedure, two modules are incorporated. SAT module is inserted in the decoder to learn the fine-grained transfer paradigm between $H_A$ and $H_B$, while LGR module conducts reasoning over a graph representation, where each node is gathered from the features in the coordinate space to represent a specific body part.

\subsection{Semantic-aware Geometric Transformation}
\begin{figure}
  \begin{center}
    \includegraphics[width=1\linewidth]{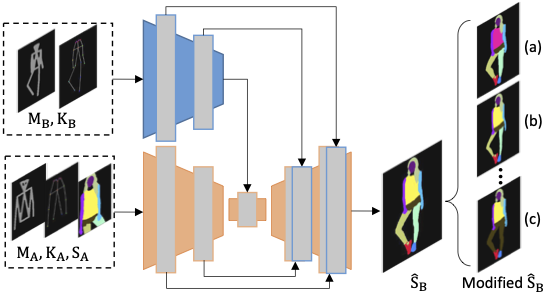}
  \end{center}
  \caption{Illustration of semantic-aware geometric transfer.}
  \vspace{-2mm}
  \label{fig:SMTN}
\end{figure}

The semantic-aware geometric transformation network is shown in Fig.~\ref{fig:SMTN}, which aims to learn the mapping $S_A \rightarrow S_B$ for producing $\hat{S}_B$ conditioned on $K_B$. The synthesized $\hat{S}_B$ contains part-level semantic labels, thus the coarse spatial mapping between corresponding human parts is provided by the $S_A \to \hat{S}_B$.
During inference, $\hat{S}_B$ can be further modified by user according to their demands for changing the body figure and clothing simultaneously (right most in Fig.~\ref{fig:SMTN}). More specifically, the clothing category in different body part region can be changed independently, and our network allows user to directly revise the shape of $\hat{S}_B$ to interactively control the body figure. 
For example, in the Fig.~\ref{fig:SMTN}(a), the $\hat{S}_B$ is slightly revised to be fatter, and the upper-cloth is changed to coat. Besides, we employ another encoder to encode the $\{\hat{S}_B, K_B\}$ to multi-scale features as the target signals (blue blocks in Fig.~\ref{fig:framework}(a)), which are inserted to the corresponding layers in the decoder of appearance transformation network. In addition, $\hat{S}_B$ is leveraged by SAT and LGR to conduct intra/inter-part transfer. Therefore, the modified $\hat{S}_B$ guides the appearance transformation with semantic-level spatial constraints to generate the portrait photo bearing desired characteristics.

\subsection{Semantic-aware Attentive Transfer}
As shown in Fig.~\ref{fig:framework}, $\Phi_l \in \mathbb{R}^{C\times H \times W}$ denotes the encoder feature of the l-th layer, and $\Phi_{L-l} \in \mathbb{R}^{C\times H \times W}$ is the corresponding decoder feature of the (L-l)-th layer. Our proposed SAT module takes the $\{\Phi_l,\Phi_{L-l},S_A,\hat{S}_B\}$ as input and generates a transformed feature $\hat{\Phi}_{L-l} \in \mathbb{R}^{C\times H \times W}$. Then $\hat{\Phi}_{L-l}$ are concatenated with $\Phi_{L-l}$ for the following operations. Specifically, $\hat{\Phi}_{L-l}=\sum_{c=1}^N \hat{\Phi}_{c,L-l}$, where c is the c-th human part category and N is total number of categories (20 for DeepFashion and 7 for Market-1501). Here we compute $\hat{\Phi}_{c,L-l}$ for each parts c respectively and add them together to form the final result. $\hat{\Phi}_{c,L-l}$ is obtained by computing the weighted mean of all locations of the c-th part in source feature $\Phi_l$:
\begin{equation}
\hat{\Phi}_{c,L-l}^{j}=\sum_{i \in S_A^c} W^{j,i} \Phi_l^i, \quad j \in \hat{S}_B^{c}.
\label{con:eq}
\end{equation}
Here $i \in {S}_A^{c}$ means the i-th location in the specific c-th part region of $S_A$, and $j \in \hat{S}_B^{c}$ is the j-th location in the same c-th part region of $\hat{S}_B$ (shown in Fig.~\ref{fig:framework}). Let $\Phi_l^i$ represent the n-dim vector extracted from the i-th location in $\Phi_l$. Then, for generating features of $\hat{\Phi}_{c,L-l}$ in location j, we employ a specific weighted matrix $W^j$ to transfer all the source information of the same part c from $\Phi_l$ via an attention mechanism. With the semantic-level spatial constraints provided by $S_A, \hat{S}_B$, SAT can directly learn the mapping of the corresponding part between source and target human images without being affected by other irrelevant regions. We assume the ${\Phi}_{c,L-l}$ is a coarse feature that is already aligned to the target pose distribution since a simple U-Net can roughly solve the pose transfer task \cite{ma2017pose}. The attention map can be calculated by
\begin{equation}
  W^{j,i}=\frac{exp(s_{i,j})}{\sum_{i \in S_A^c} exp(s_{i,j})}, \quad s_{i,j}=g(\Phi_{L-l}^j)^T \theta(\Phi_l^i).
  \label{con:matrix}
\end{equation}
Here, the $g(\cdot)$ and $\theta(\cdot)$ are two mapping functions that transfer the n-dim vectors $\Phi_{L-l}^j$ and $\Phi_l^i$ to the same latent space. We conduct a dot-product operation to measure the similarity between the feature of $g(\Phi_{L-l})$ in location j and that of $\theta(\Phi_{l})$ in location i. 

In this way, SAT directly transfer feature from source to target location for each part through an attention mechanism, which is actually an intra-part transfer. Besides, it well preserves the intrinsic appearance structure by capturing long-range dependencies among all pixels in relevant region. For example, when synthesizing eyes region, the eyes features in source portrait are transferred here to guide the generation process, and attributes of whole face region are also properly considered since orientation and shape of eyes also implicitly depend on them.

\subsection{Layout Graph Reasoning}
The goal of LGR module is to strengthen the model ability of inferring invisible body part and guaranteeing their global coherency, which is achieved via an inter-part reasoning mechanism. For example, to infer the leg region invisible in the source portrait, the skin color of the inferred leg should be consistent with that of the visible skin. 

We employ body skeleton to depict distinct correlations between different parts. which is represented as an adjacency matrix $E \in \{0,1\}^{N\times N}$ showing the connections of all the human parts such as hair and face that are adjacent. We construct an undirected graph $G=(V,E)$, where V denotes the human part node. The input of LGR is $\{G,\hat{S}_B,\Phi_{L-l}\}$ and the output is $\tilde{\Phi}_{L-l} \in \mathbb{R}^{C\times H \times W}$. As shown in Fig.~\ref{fig:framework}, the LGR is constituted by three steps: 1) gathering the features from coordinate space to graph space; 2) propagating graph representation via Graph Convolution~\cite{Kipf2016SemiSupervisedCW}; 3) distributing graph representation back to coordinate space. \newline
\textbf{\emph{Step 1)}} We compute the high-level graph representation $Q \in \mathbb{R}^{N \times D}$ of all $N$ nodes, where $D$ is the dimension for each node. First, we adopt a mapping function $P(\cdot)$ that transfers the decoder feature $\Phi_{L-r}$ of the (L-r)-th layer from coordinate space to graph space as
\begin{equation} 
Q_p=P(\Phi_{L-r},W),
\label{con:mapping}
\end{equation}
where $W$ is a set of trainable parameters of the mapping function $P(\cdot)$ and $Q_p \in \mathbb{R}^{N\times D}$ denotes the projected graph feature that is a basic representation with the same dimension of $Q$. Second, we directly sample the feature $\Phi_{L-r}$ from coordinate space to graph space with the guidance of semantic information $\hat{S}_B$, which lies in the same coordinate space with $\Phi_{L-r}$. Then the sampled graph feature $Q_s \in \mathbb{R}^{N\times D}$ can be formalized as
\begin{equation}
Q_s=S(\Phi_{L-r},\hat{S}_B),
\label{con:sampling}
\end{equation}
where $S(\cdot)$ represents the sampling operation. For face region as an example, we employ global average pooling for all features in the face region to obtain the graph feature of face node. Generalized, the c-th human part node feature $Q_s^c \in \mathbb{R}^{1\times D}$ can be formalized as
\begin{equation}
Q_s^c=GlobalAvgPool(\Phi_{L-r}^j,j\in \hat{S}_B^c).
\label{con:sampling2}
\end{equation}
Then a convolution layer with a nonlinear function $\sigma [\cdot]$ is adopted here to process the concatenated graph feature $[Q_p,Q_s]$ from $N\times 2D$-dim to $N\times D$-dim:
\begin{equation}
Q=\sigma [Conv(Q_p,Q_s)].
\label{con:Q}
\end{equation}
Each node feature of graph $Q$ represents a specific semantic part (e.g., face) since the visual features that are correlated to the part are directly aggregated via mapping and semantic-aware sampling to depict the characteristic of its corresponding node. Capturing relations between different parts are now simplified to capturing the relations between the features of corresponding nodes. \newline
\textbf{\emph{Step 2)}} We then propagate the graph features over all part nodes with matrix multiplication:
\begin{equation}
Q'=\sigma'(EQW^e),
\label{con:gcn}
\end{equation}
where $W^e \in \mathbb{R}^{D\times D}$ is a trainable weight matrix, and $E$ is the mentioned hierarchical knowledge represented as a symmetric adjacency matrix for propagating node information. We conduct such a propagation procedure multiple times (n times for ablation study). \newline
\textbf{\emph{Step 3)}} Then an inverse mapping function with trainable weight $W'\in \mathbb{R}^{N\times N}$ is introduced to project the graph representation back to the coordinate space to obtain $\tilde{\Phi}_{L-r}=P'(Q,W')$, where $\tilde{\Phi}_{L-r}$ is concatenated with $\Phi_{L-r}$ for the following operations.

Normal convolution operations are effective at capturing local relations, but they are typically inefficient at modeling global relations among distant regions, especially when changing human poses where we need to mine the relations among distant specific parts. Through a skeleton graph, the reasoning procedure of LGR enables high-level information to propagate among body parts, which essentially provides more direct interaction among different body regions.

\subsection{Discriminator and Objective Function}
The goal of discriminator is to constrain generator for synthesizing realistic portrait in an adversarial training manner. Specifically, the generated portrait needs to preserve the identity and intrinsic structure of the source person, while satisfying desired pose. Therefore, we introduce there lightweight discriminators $D_1 \sim D_3$ with specific loss functions to meet these requirements respectively.\newline
\textbf{Cross Entropy Loss.} To encourage the semantic-aware geometric transformation part of our model to generate high quality $\hat{S}_B$ to guide the following appearance transformation, we adopt a per-pixel cross-entropy loss $L_s$ for $\hat{S}_B$ and ground truth labels $S_B$, which is usually used in semantic segmentation tasks.
\begin{equation}
  L_s=-\left \| S_B \odot \log(\hat{S}_B) \right \|_1,
\label{con:CE}
\end{equation}
where the $\odot$ denotes the element-wise multiplication.\newline 
\textbf{Adversarial Losses.} 
For space limit, we define the adversarial loss for generator and discriminator respectively:
\begin{equation}
\begin{aligned}
L_{D}(D,\hat{H},H,C)&=-\mathbb{E}_{(H,C)\sim p_{data}}[\log D(H\mid C)] \\
&-\mathbb{E}_{C\sim p_{data},\hat{H}\sim p_g}[\log(1-D(\hat{H}\mid C))],
\end{aligned}
\label{con:D_adv}
\end{equation}
\begin{equation}
L_{G}(D,\hat{H},C)=\mathbb{E}_{C\sim p_{data},\hat{H}\sim p_g}[\log(1-D(\hat{H}\mid C))],
\label{con:D_adv}
\end{equation}
where D is the discriminator and $\hat{H}$ is generated by the generator. $C$ is the conditional information, $p_{data}$ denotes the true data distribution and $p_g$ means the generator's distribution.
The adversarial losses for geometric transformation are
\begin{equation}
L_D^S=L_D(D_1,\hat{S}_B,S_B,K_B),
\label{con:l_s_d}
\end{equation}
\begin{equation}
L_G^S=L_G(D_1,\hat{S}_B,K_B),
\label{con:l_s_g}
\end{equation}
where $D_1$ is applied to judge whether $\hat{S}_B$ lies in true data distribution and whether it is consistent with the target pose $K_B$.

Discriminators $D_2,D_3$ are adopted to constrain the posture consistency and the identity preserving of the generated human image $\hat{H}_b$, respectively. Formally,
\begin{equation}
\begin{aligned}
L_D^H=L_D(D_2,\hat{H}_B,H_B,K_B)+L_D(D_3,\hat{H}_B,H_B,H_A),
\end{aligned}
\label{con:d2}
\end{equation}
\begin{equation}
\begin{aligned}
L_G^H&=L_G(D_2,\hat{H}_B,K_B)+L_G(D_3,\hat{H}_B,H_A),
\end{aligned}
\label{con:d2}
\end{equation}
where $K_B$ and $H_A$ denote the condition information of the discriminators to constrain the pose and identity. \newline
\textbf{$\bf{L_1}$ Losses.}
Since we have paired data, we compute the pixel-wise $L_1$ loss between ground truth image and generated image as
\begin{equation}
  L_1=\left \| \hat{H}_B - H_B \right \|_1.
\label{con:L1}
\end{equation}
We also integrate a perceptual loss $L_{p}$ to improve the visual quality of the generated results:
\begin{equation}
    L_p=\left \| \nu(\hat{H}_B) - \nu(H_B) \right \|_1.
\label{con:perceptual}
\end{equation}
where $\nu(H)$ is the feature map of image $H$ of $conv1\_2$ layer in VGG-19~\cite{simonyan2014very} model pre-trained on ImageNet.

Therefore, the final objective is computed as follows:
\begin{equation}
L_D=\lambda_1 L_D^S+ \lambda_2 L_D^H
\label{con:final_d}
\end{equation}
\begin{equation}
L_G=\lambda_1 L_G^S+\lambda_2 L_D^H+ \lambda_3 (L_1+L_p)+ \lambda_4 L_s
\label{con:final_g}
\end{equation}
where $\lambda_i (i=1,\cdot\cdot\cdot,4)$ is a weighting factor that controls the relative importance of each term.
\section{Experiments} 
\subsection{Experimental Setting}
To our best knowledge, there is no method that targets at all of our goals at the same time, which means we do not have proper baselines to compare with. Therefore, we first compare our method with previous pose transfer methods in Sec.~\ref{Compared} and Sec.~\ref{ablation_study}, which is one of our sub tasks. Then we provide other recapture results from Sec.~\ref{other1} to Sec.~\ref{other3} to further illustrate the effectiveness and superiority of our method.\newline
\textbf{Datasets.} We use DeepFashion~\cite{liu2016deepfashion} and Market-1501~\cite{zheng2015scalable} datasets in experiments. DeepFashion contains 52,712 person images of size 256 $\times$ 256. We split it to training/test following the settings in~\cite{ma2017pose,Siarohin2018DeformableGF} and select 89,262 pairs for training and 12,000 pairs for testing, where each pair includes the same person in different pose. Market-1501 contains 322,668 images of size 128 $\times$ 64 collected from 1,501 persons. Following the setting in~\cite{Siarohin2018DeformableGF,zhu2019progressive}, we collect 263,632 training pairs and randomly sample 12,000 pairs for testing. \newline
\textbf{Evaluation.} We follow the evaluation setting of~\cite{ma2017pose,Siarohin2018DeformableGF,pumarola2018unsupervised,ma2018disentangled,song2019unsupervised,li2019dense,zhu2019progressive}. Inception Score (IS, higher is better)~\cite{Salimans2016ImprovedTF} and Structure Similarity (SSIM, higher is better)~\cite{wang2004image} are adopted for evaluating generation quality and consistency. We also adopt M-IS and M-SSIM, where background is ignored. We additionally use Fréchet Inception Distance (FID, lower is better)~\cite{Heusel2017GANsTB} to calculate the distance between true image distribution and that of generated ones. \newline
\textbf{Implementation Details.} We adopt Adam optimizer~\cite{kingma2014adam} in all experiments with hyper-parameter $\{\alpha=0.0002, \beta_1=0.5, \beta_2=0.999\}$. The weighting factors of the full objective are set to $\{\lambda_1=1,\lambda_2=5,\lambda_3=10,\lambda_4=50\}$ for Market-1501 and $\{\lambda_1=1,\lambda_2=5,\lambda_3=2,\lambda_4=50\}$ for DeepFashion. We apply Spectral Norm~\cite{Miyato2018SpectralNF} to all the layers in both generator and discriminator. The batch size is set to 4 for DeepFashion and 20 for Market-1501. Besides, the number of human semantic part category is set to 20 for DeepFashion and 7 for Market-1501 due to the different image resolution.
\begin{figure}[b]
  \vspace{-2mm}
  \begin{center}
  \includegraphics[width=1\linewidth]{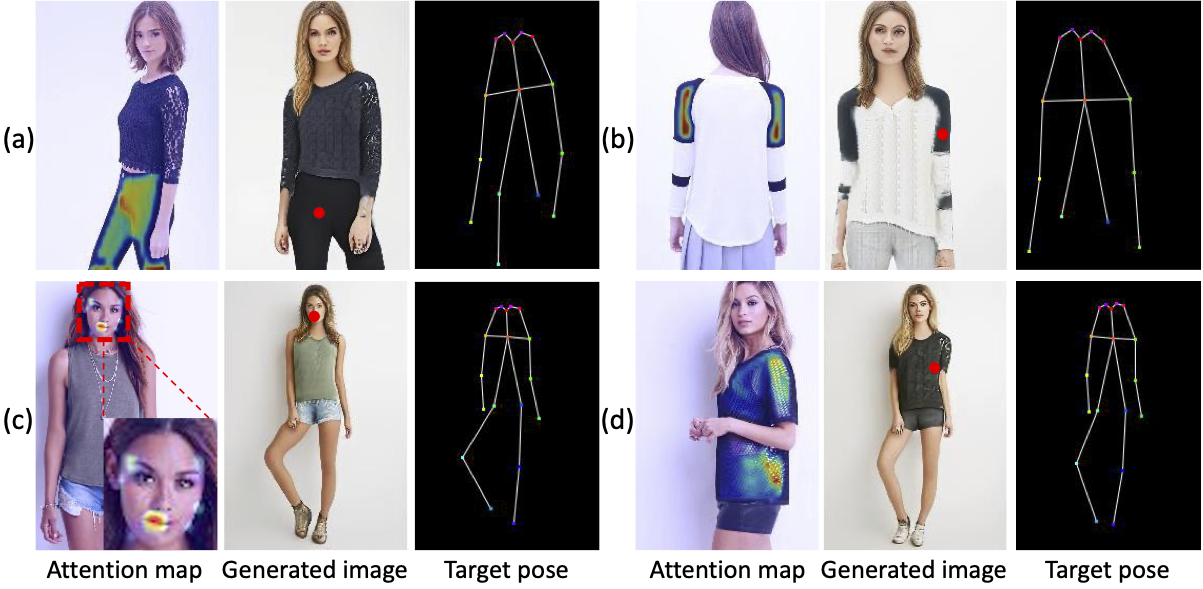}
\end{center}
\vspace{-2mm}
  \caption{Attention map visualization. The attention map shows the attentive region in generating a specific pixel (denoted as red circle), which illustrates the effectiveness and interpretability of the proposed SAT module.}
  \label{fig:visualization}
\end{figure}

\subsection{Compared with State-of-the-Art Methods}
\label{Compared}
We compare with state-of-the-art pose transfer methods including PG\textsuperscript{2}~\cite{ma2017pose}, Def-GAN~\cite{Siarohin2018DeformableGF}, UPIS~\cite{pumarola2018unsupervised}, DPIG~\cite{ma2018disentangled}, UPIG-SPT~\cite{song2019unsupervised}, DIAF~\cite{li2019dense}, PATN~\cite{zhu2019progressive} on DeepFashion and Market-1501 datasets.\newline
\textbf{Quantitative Comparison.} The IS and SSIM results are shown in Tab.~\ref{tb:SOTA}, and the FID results are shown in Tab.~\ref{tb:FID}. Note, we also directly use their released well-trained models over our test pairs for fair comparison (denoted as * in Tab.~\ref{tb:SOTA}) since their original testing pair lists are not available. Our method outperforms all state-of-the-arts, although some images in our testing set include part of their training images. Our method is the best w.r.t. SSIM metric on both datasets, meaning our generated results are more consistent with source images in both structure and appearance. Our method also achieve the best IS and FID on DeepFashion and best M-IS and FID scores on Market-1501, which prove its good generalizability on real data distribution.   

In addition, previous methods do not adopt parsing maps for training and testing since their goal is not recapturing. Therefore, we conduct some extra experiments (Tab.~\ref{tb:parsing}) on a representative approach PATN~\cite{zhu2019progressive} to explore the influences of its using extra source/target parsing maps during training/test phase. Although the performance of PATN slightly improves when utilizing parsing maps, our results are still the best. Note that using ground truth (target) parsing for testing, our performance improves significantly, which means the boosted geometric transformation of our method will further improve final results. These experiments reveal the effectiveness and superiority of our proposed transformation paradigm and two modules (SAT and LGR).
\begin{table}
  \begin{center}
  \begin{tabular}{lcc|cccc}
  \toprule
  \multirow{2}{*}{Methods} & \multicolumn{2}{c|}{DeepFashion} & \multicolumn{4}{c}{Market-1501}     \\ \cmidrule{2-7} 
                                  & IS     & SSIM         & IS     & M-IS  & SSIM   & M-SSIM \\ \midrule
  PG\textsuperscript{2}           & 3.090  & 0.762        & 3.460  & 3.435    & 0.253  & 0.792          \\
  UPIS                            & 2.970  & 0.747        & $-$    & $-$      & $-$    & $-$         \\
  DPIG                            & 3.228  & 0.614        & 3.483  & 3.491    & 0.099  & 0.614      \\ 
  Def-GAN                         & 3.439  & 0.756        & 3.185  & 3.502    & 0.290  & 0.805    \\
  UPIG-SPT                        & 3.441  & 0.736        & 3.499  & 3.680    & 0.203  & 0.758   \\
  DIAF                            & 3.338  & 0.778        & 3.010  & 3.700    & 0.308  & 0.874     \\ 
  PATN                            & 3.209  & 0.773        & 3.323  & 3.773    & 0.311  & 0.811      \\ \midrule
  PG\textsuperscript{2}*          & 3.294  & 0.762        & \textbf{3.382}  & 3.384    & 0.259  & 0.780      \\
  Def-GAN*                        & 3.327  & 0.763        & 3.188  & 3.516    & 0.289  & 0.782     \\
  PATN*                           & 3.214  & 0.774        & 3.256  & 3.741    & 0.291  & 0.784     \\
  Ours                            & \textbf{3.528}  & \textbf{0.778} & 3.275  & \textbf{3.776} & \textbf{0.295} & \textbf{0.793} \\ \bottomrule
\end{tabular}
\end{center}
\caption{Comparison with state-of-the-art methods.}
\label{tb:SOTA}
\vspace{-2mm}
\end{table}
\begin{table}
  \vspace{-2mm}
  \begin{center}
  \begin{tabular}{lcccc}
  \toprule
  Methods & PG\textsuperscript{2}\cite{ma2017pose} & Def-GAN\cite{Siarohin2018DeformableGF} & PATN\cite{zhu2019progressive} & Ours  \\  \midrule
  DeepFashion &41.333     &22.664       &19.545      &\textbf{15.936}  \\ 
  Market-1501 &127.644    &74.279       &68.132      &\textbf{40.458}  \\ \bottomrule
\end{tabular}
\end{center}
\caption{FID score of different methods.}
\label{tb:FID}
\vspace{-2mm}
\end{table}
\begin{table}
  \vspace{-2mm}
  \begin{center}
  \begin{tabular}{lccccc}
  \toprule
  \multirow{2}{*}{Methods} & \multicolumn{2}{c}{Parsing Map}  & \multicolumn{3}{c}{DeepFashion} \\ \cmidrule{4-6}
                                   & Source     & Target      & IS     & SSIM   & FID           \\ \midrule
  PATN~\cite{zhu2019progressive}   &            &             & 3.214  & 0.774  & 19.545      \\\midrule
  PATN~\cite{zhu2019progressive}   &\checkmark  &             & 3.277  & 0.761  & 20.827         \\
  Ours                             &\checkmark  &             & \textbf{3.528}  & \textbf{0.778}  & \textbf{15.936}        \\ \midrule
  PATN~\cite{zhu2019progressive}   &\checkmark  &\checkmark   & 3.682  & 0.792  & 13.973        \\
  Ours                             &\checkmark  &\checkmark   & \textbf{3.855}  & \textbf{0.813}  & \textbf{10.431}         \\ \bottomrule
\end{tabular}
\end{center}
\caption{Exploring the influence of utilizing parsing maps.}
\label{tb:parsing}
\vspace{-6mm}
\end{table}
\begin{figure*}
  \begin{center}
  \includegraphics[width=0.96\linewidth]{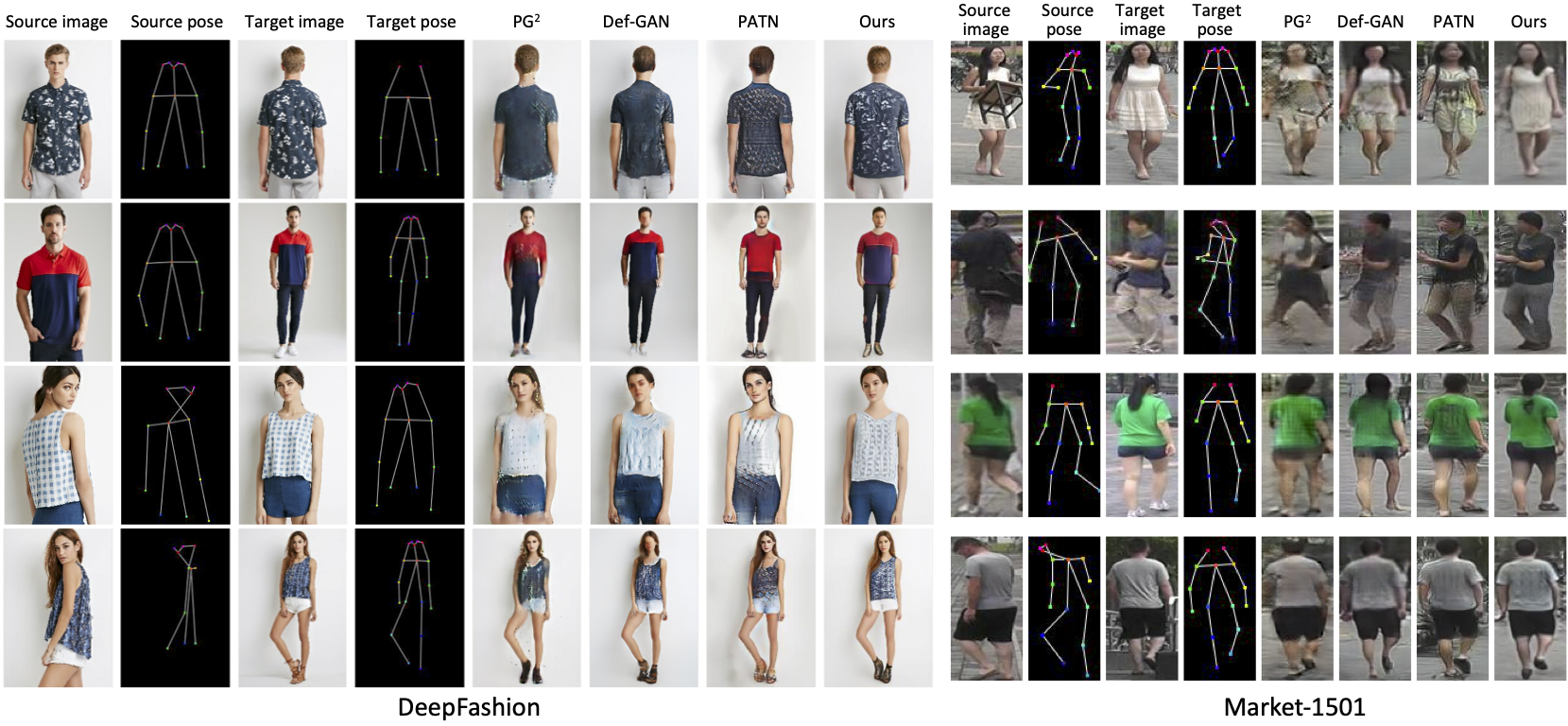}
\end{center}
\vspace{-3mm}
  \caption{Qualitative comparisons with state-of-the-art methods. It can be observed that our method effectively preserves the intrinsic appearance structure of source human images and also texture details. Please zoom in to see details.}
  \label{fig:qualitative}
\end{figure*}
\begin{table*}
  \begin{center}
     \begin{tabular}{lccc|ccccc}
     \toprule
     \multirow{4}{*}{Methods} & \multicolumn{3}{c|}{DeepFashion} & \multicolumn{5}{c}{Market-1501}     \\ \cmidrule{2-9} 
                       & IS    & SSIM  & FID     & IS     & Mask-IS  & SSIM  &Mask-SSIM  &FID\\ \midrule
     Basic             & 3.392 & 0.763 & 21.443  & 2.991  & 3.649 & 0.285 & 0.793 & 57.162\\
     Basic+SAT($l+2$)    & 3.441 & 0.761 & 16.842        & 3.049 & 3.627 & 0.273 & 0.791 & 51.362  \\
     Basic+SAT($l+1$)    & 3.512 & 0.766 & 20.061  & 3.027 & 3.619 & 0.269 & 0.786 & 53.021   \\ 
     Basic+SAT($l$)    & 3.416 & 0.776 & 16.841  & 3.111 & 3.652 & 0.276 & \textbf{0.798} & 53.019  \\ \midrule
     Basic+SAT($l$)+LGR($r+1,n$)    & \textbf{3.528} & \textbf{0.778} & 15.936  & \textbf{3.275} & \textbf{3.776} & \textbf{0.295} & 0.793 & \textbf{40.458} \\    
     Basic+SAT($l$)+LGR($r+1,n+1$)    & 3.417 & 0.773   & 20.556 & 3.149 & 3.668 & 0.273 & 0.797 & 48.372      \\
     Basic+SAT($l$)+LGR($r,n$) & 3.349  & 0.769  & \textbf{14.879} & 3.127 & 3.667   & 0.278 & 0.795 & 51.247     \\ \bottomrule
     \end{tabular}
  \end{center}
     \caption{Quantitative results of ablation study. Effectiveness of our SAT and LGR modules is well proved. `n' indicates that we conduct the propagation procedure of graph n times in LGR. Note that, considering the different resolution between two datasets, $l$ is set to 3 for DeepFashion and 4 for Market-1501. The $r$ and $n$ are set to 5 and 2 for both datasets.}
     \label{tb:Comparison ave scores}
     \vspace{-5mm}
  \end{table*} \newline 
\textbf{Qualitative Comparison.} All qualitative comparisons on DeepFashion and Market-1501 are shown in Fig.~\ref{fig:qualitative}. It can be observed that our method effectively preserves texture details and intrinsic appearance structure of source human. For example, in the first row, the pattern structure on the upper-cloth is effectively transferred while all compared methods give blurry and coarse results, which fail to correctly capture and transfer such information. The fourth row shows that our method better guarantee the global coherency of the inferred body parts. From the comparisons on Market-1501 we can see that our method achieves sharper results despite the low resolution of testing images, with the guidance of SAT module. Comparatively, baseline methods get blurry results that are even inconsistent with the source appearance.
\subsection{Component Analysis}
\label{ablation_study}
\textbf{Visualization of SAT module.} 
Visualized attention map in Fig.~\ref{fig:visualization} demonstrates effectiveness and interpretability of our method. For generating one pixel (denoted in red circle), SAT correctly locates the most relevant region in source image and learns to utilize long-range dependencies among all pixels in that region for better inference. We can observe from Fig.~\ref{fig:visualization}(b) that attention is equally allocated to both shoulders. This indicates SAT correctly captures the intrinsic appearance structure (symmetrical style) in source image and effectively transfers it for generation. From Fig.~\ref{fig:visualization}(c), we can investigate which region of the source image is attended when generating pixels of the mouth. 
Despite the large scale and angle changes between source and target images, SAT still accurately locate the mouth region in the source image, with identity better preserved considering shape of the whole face region.\newline
\begin{figure}
  \begin{center}
  \includegraphics[width=1\linewidth]{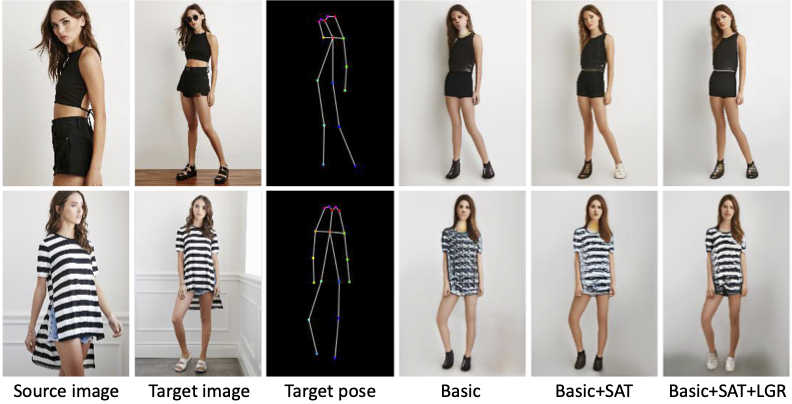}
  \end{center}
  \vspace{-1mm}
  \caption{Ablation study. SAT module more effectively preserves the intrinsic appearance structure and LGR well infers the invisible region and improves the global coherence.}
  \label{fig:ablation}
  \vspace{-3mm}
\end{figure}
\textbf{Quantitative Ablation Study.}
To investigate effectiveness of each module, we conduct ablation studies shown in Tab.~\ref{tb:Comparison ave scores}. `Basic' refers to the basic network without using SAT and LGR, which achieves inferior results. 
For dealing with the different resolution between two datasets, we set $l$ to 3 for DeepFashion and 4 for Market-1501, respectively. Besides, $r$ and $n$ are set to 5 and 2 for both datasets. In our network, the spatial size of features in the $\{l=3,l=4,l=5\}$-th layer is $\{64\times 64,32\times 32,16\times 16\}$ respectively. The influences of different hyperparameters in each module can be observed in Tab.~\ref{tb:Comparison ave scores}, which demonstrate that $\{l=3,r=6,n=2\}$ and $\{l=4,r=6,n=2\}$ are the best configurations for DeepFashion and Market-1501 respectively. These results indicate the attentive transfer process is better to be conducted on middle-level features while reasoning is better to be conducted on high-level features.
\begin{figure}
  \begin{center}
  \includegraphics[width=1\linewidth]{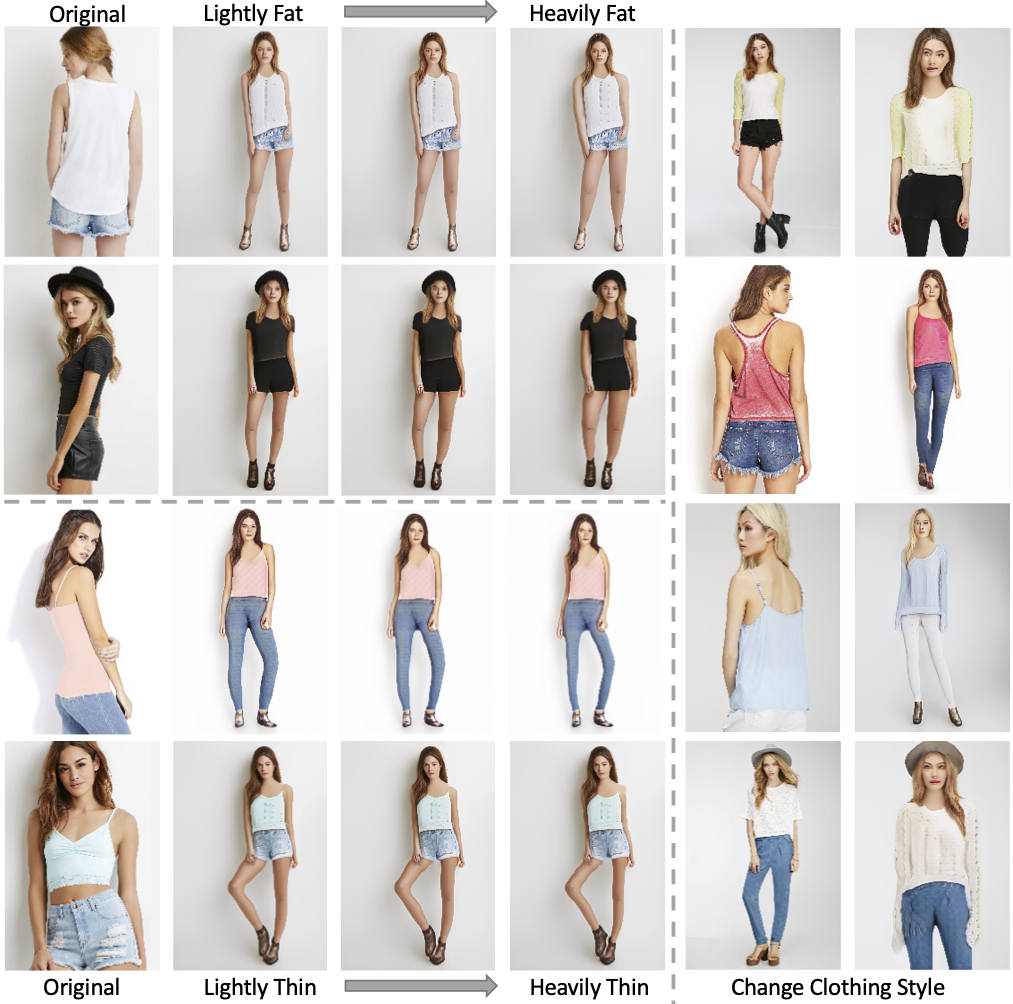}
\end{center}
\vspace{-2.5mm}
  \caption{Changing the body figure and clothing style.}
  \label{fig:cloth}
  \vspace{-3.5mm}
\end{figure}
\begin{figure}[b]
  \vspace{-2.5mm}
  \begin{center}
    \includegraphics[width=1\linewidth]{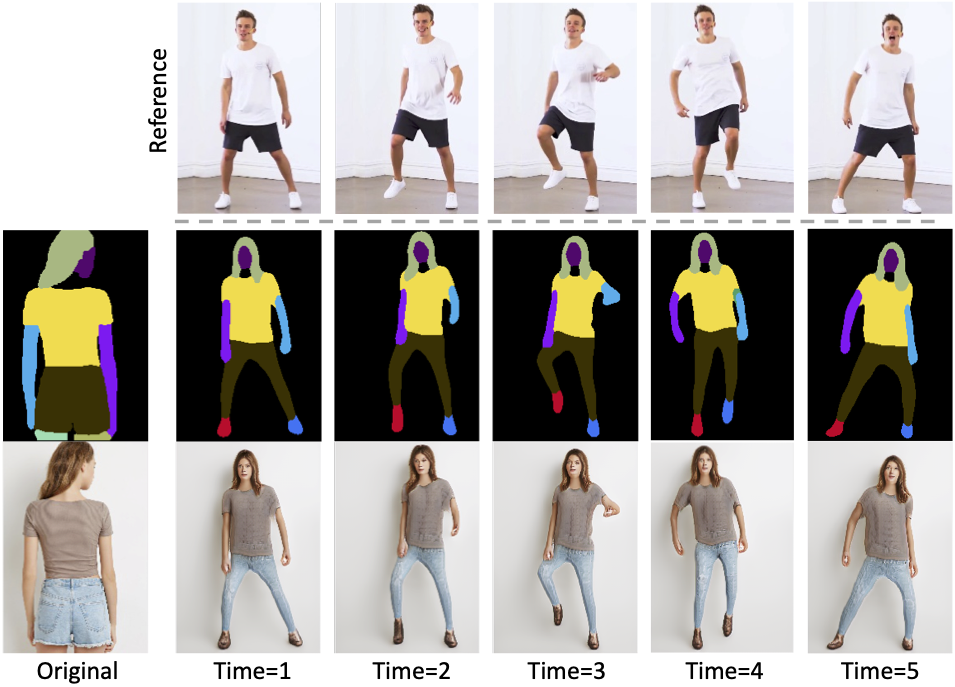}
  \end{center}
  \vspace{-2mm}
    \caption{Dancing as the reference with a long pants. More results in appendix.}
    \label{fig:video}
  \end{figure}\newline
\textbf{Qualitative Ablation Study.}
The qualitative results in Fig.~\ref{fig:ablation} show the impact of each module. As shown in the second row, the appearance structure and texture details are more consistent with the source image with the help of SAT. The first row show that LGR better infer the invisible region and guarantees the global coherence among each part, meaning it learns their intrinsic relationships. For example, two shoes usually are the same color and invisible skin color should accord with the visible region.

\begin{figure}
  \begin{center}
  \includegraphics[width=1\linewidth]{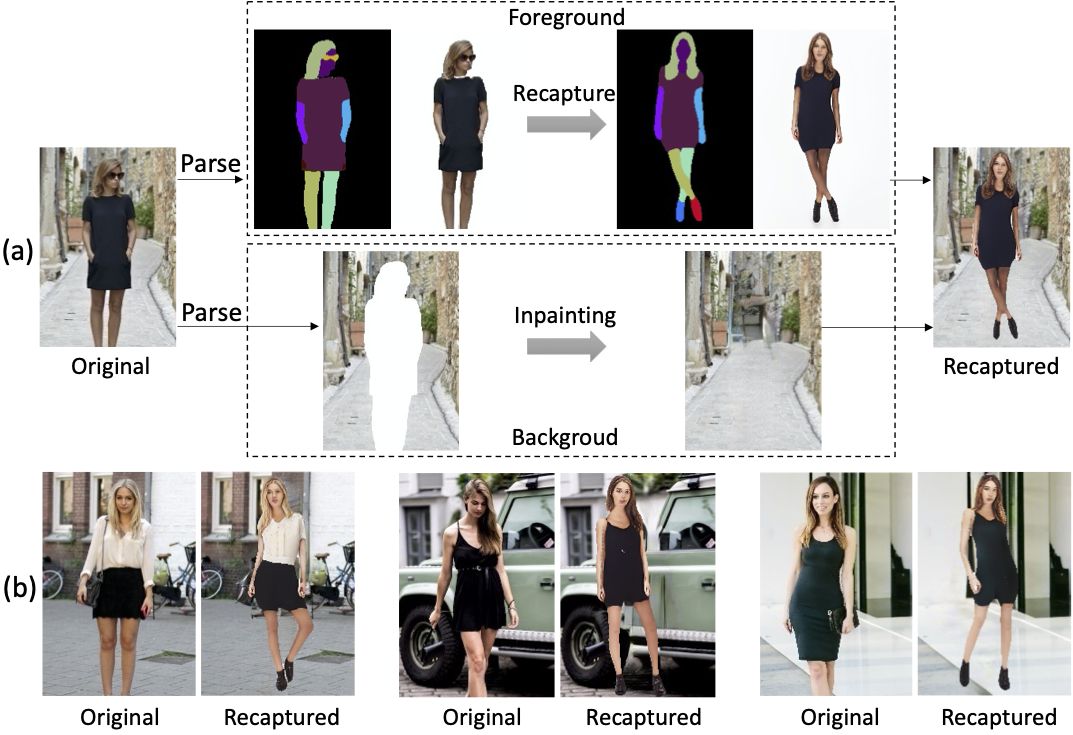}
\end{center}
\vspace{-2mm}
  \caption{Applying our method to the portraits in the wild. (a) System pipline. (b) Some results.}
  \label{fig:real}
  \vspace{-3.5mm}
\end{figure} 

\subsection{Changing Body Figure and Clothing}
\label{other1}
We show examples of changing body figure and clothing in Fig.~\ref{fig:cloth}, which are achieved through manipulating the specific body part regions in generated parsing maps. The generation process is strictly subject to the layout constraint, thus our model can control the degree of body figure changing utilizing the edited parsing map by users. It can be seen although the clothing is changed, the synthesized new clothes are visually-realistic and semantically-consistent. 

\subsection{Portrait Recapturing in the Wild}
\label{other2}
There is domain gap between training data (DeepFashion and Market-1501) and in-the-wild data in terms of light, background, etc. Hence, we apply some extra data processing to achieve portrait recapturing in the wild, with pipeline shown in Fig.~\ref{fig:real} (a).  
First, we parse the portrait to extract foreground and background images, and then apply our model to the foreground human and obtain the desired human image. Second, we utilize the off-the-shelf inpainting model~\cite{yu2019free} to fill these missing regions in the background image. Finally, we combine the foreground human and background to generate the desired portrait.
The results in the wild (Fig.~\ref{fig:real}(b)) further demonstrate the effectiveness and practicability of our portrait recapture model.

\subsection{Video Recapturing}
\label{other3}
In some scenarios, users may want to recapture their portraits forming a short video. This is a challenging goal since extra difficulties need be considered, such as consistency of human appearance and posture in the time sequence. To examine effectiveness of our method in this setting, we simply perform our method on every frame to produce a video, as shown in Fig.~\ref{fig:video}. It can be seen the results are fairly nice and stable, which further demonstrate robustness and practicability of our model. More video results are provided in appendix and \href{https://youtu.be/vTyq9HL6jgw}{\color{magenta}{https://youtu.be/vTyq9HL6jgw}}.

\section{Conclusion}
In this paper, we propose a portrait photo recapture system that allows users to easily edit their portraits to desired posture/view, body figure and clothing. The semantic-aware geometric and appearance transformation with two novel modules (SAT $\&$ LGR) are performed to solve the unique challenges in the task. SAT module is designed for intra-part transfer to deal with non-rigid deformation of human body and preserve intrinsic appearance structure. LGR module is devised for inter-part reasoning to infer invisible body parts and guarantee global coherence of generated human body parts. Extensive qualitative and quantitative experiments demonstrate the effectiveness and practicability of our method. 


\bibliographystyle{ACM-Reference-Format}
\bibliography{sample-base}
\newpage
\appendix

\section{Appendix}

\subsection{Comparisons with State-of-the-Arts}
\begin{figure*}
  \begin{center}
    \includegraphics[width=1\linewidth]{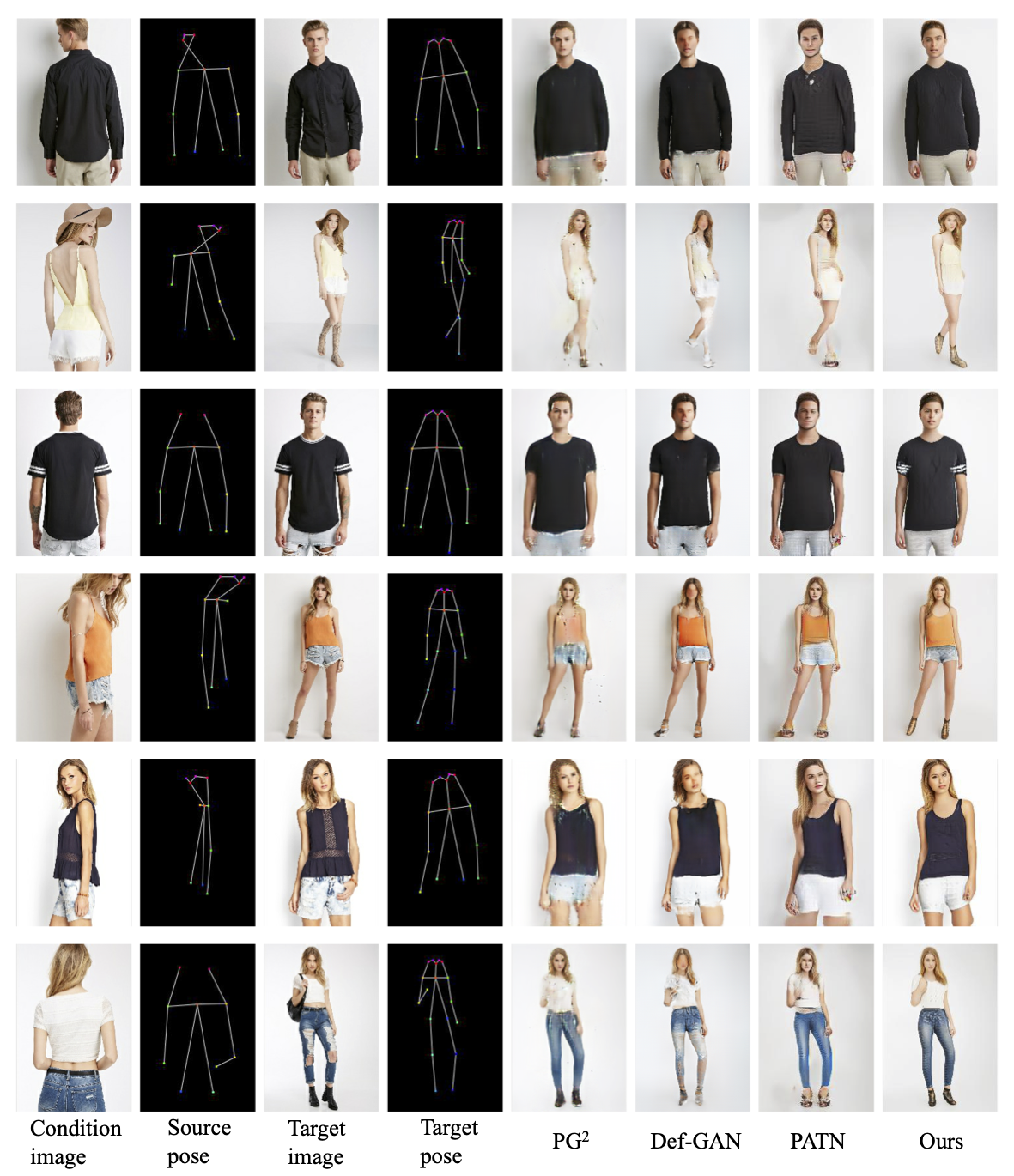}
  \end{center}
  \caption{Qualitative comparisons with state-of-the-art methods on DeepFashion.}
  \label{fig:sota}
\end{figure*}

In Fig.~\ref{fig:sota}, we show more results compared with state-of-the-art pose transfer methods on DeepFashion~\cite{liu2016deepfashion}, including PG\textsuperscript{2}~\cite{ma2017pose}, Def-GAN~\cite{Siarohin2018DeformableGF} and PATN~\cite{zhu2019progressive}. We directly adopt their released code and well-trained model for comparison.
It can be observed that our method more effectively preserves the intrinsic appearance structure and accurate texture details of the source human image. The invisible regions of human body are properly inferred with better consistency in our results. Besides, our method also generates more realistic faces and better maintains the facial identity.

\subsection{Recapturing Results}
\begin{figure*}
  \begin{center}
    \includegraphics[width=0.9\linewidth]{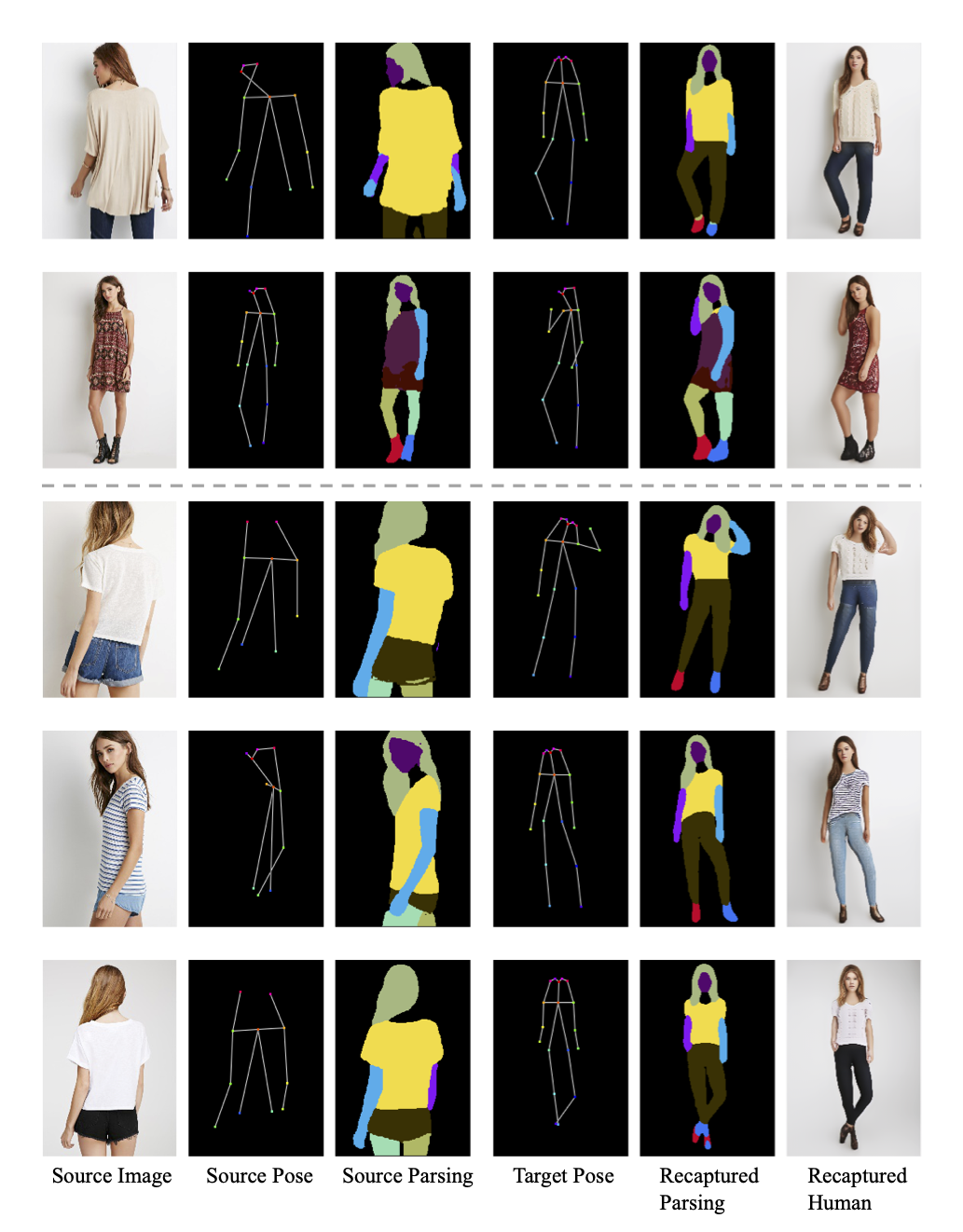}
  \end{center}
  \caption{The first two rows show the results of changing posture and figure (fatter) simultaneously. The last three rows show the results of changing posture, figure (fatter) and clothing simultaneously.}
  \label{fig:fat}
\end{figure*}

\begin{figure*}
  \begin{center}
    \includegraphics[width=0.9\linewidth]{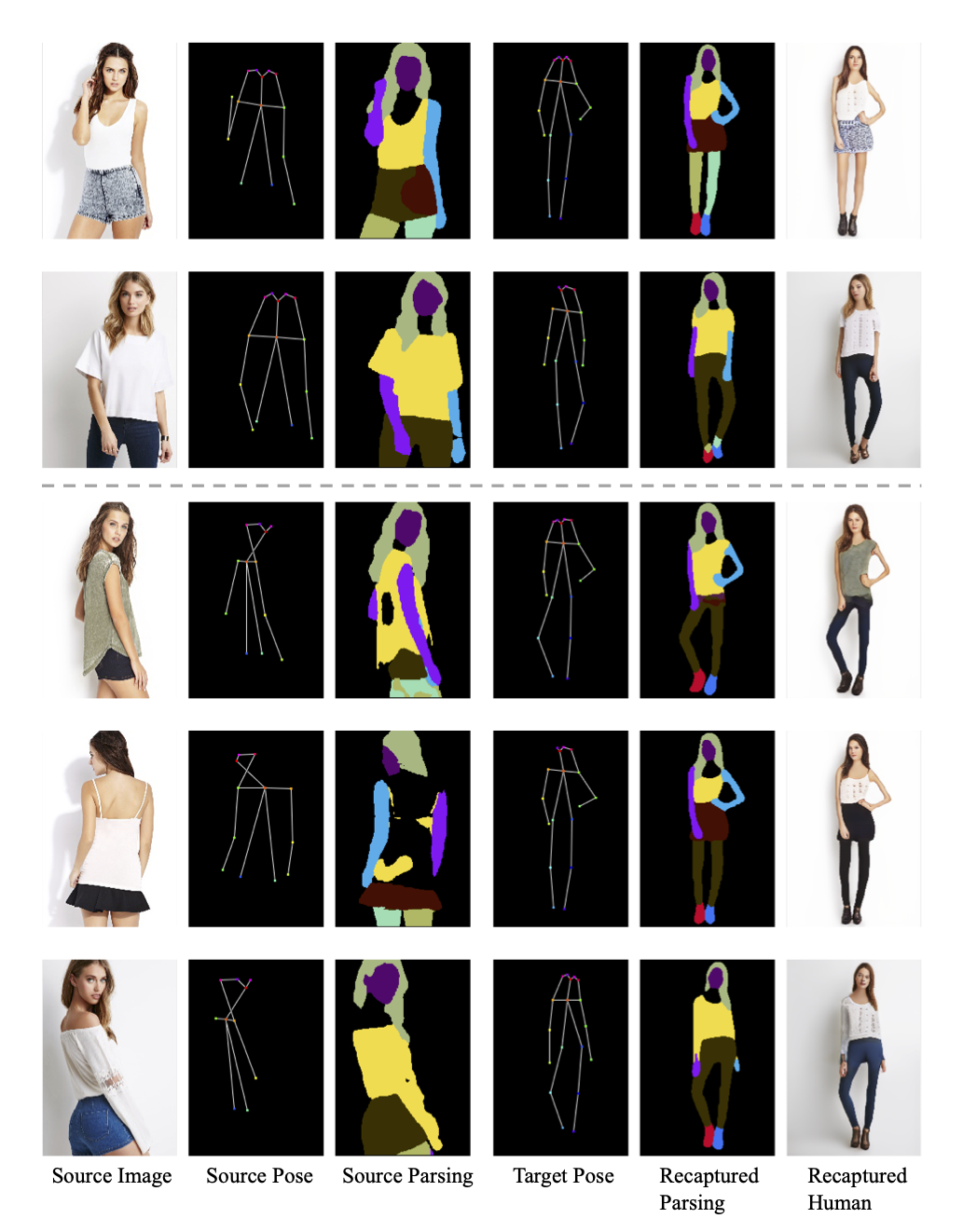}
  \end{center}
  \caption{The first two rows show the results of changing posture and figure (thinner) simultaneously. The last three rows show the results of changing posture, figure (thinner) and clothing simultaneously.}
  \label{fig:thin}
\end{figure*}

In Fig.~\ref{fig:fat} and Fig.~\ref{fig:thin}, we present more portrait recapturing results of our system. The first two rows in Fig.~\ref{fig:fat} and Fig.~\ref{fig:thin} show the results of changing posture and body stature simultaneously. The last three rows show the results of changing posture, body figure and clothing simultaneously. 

These results are all visually realistic and semantically consistent, which further demonstrate the effectiveness and practicality of our system.
In addition, from the fourth row in Fig~\ref{fig:thin}, we can observe that our system is robust to bad source parsing and produces better transformed parsing for the semantic-aware appearance transformation. The model robustness is importance since the state-of-the-art human parser may produce poor parsing results.  

\subsection{Video Recapturing Results}
To further expand the application scenarios of our system, we examine the effectiveness of our model in video recapturing. The goal of video recapturing is to generate a short video according to a given portrait image and a reference video, where the movement in the generated video is required be consistent with the reference video.  
Since extra issues need to be properly considered, such as the consistency of human appearance and posture in the time sequence, it is a more challenging task.

We complete such a task through three steps. First, we split the reference video into a sequence of frames, and then adopt HPE~\cite{cao2017realtime} to estimate the reference pose of each frame. Second, we simply apply our recapture model to the given portrait and each reference pose, forming a sequence of recaptured human images. Note, the identity of the generated human is the same with the one in the given portrait, and the body figure and clothing can be modified through our model. Third, we stack all the generated human images in the order of reference frames and synthesize them into a video, with the same FPS as the reference video. 

The results are shown in Fig.~\ref{fig:video}, and more video results are provided at \href{https://youtu.be/vTyq9HL6jgw}{\color{black}{https://youtu.be/vTyq9HL6jgw}}. Although we simply applying our model frame by frame without using any smoothing technique, the results are still nice and stable, which demonstrate the effectiveness and capability of our model in maintaining the appearance and shape consistency.

\begin{figure*}
  \begin{center}
    \includegraphics[width=0.88\linewidth]{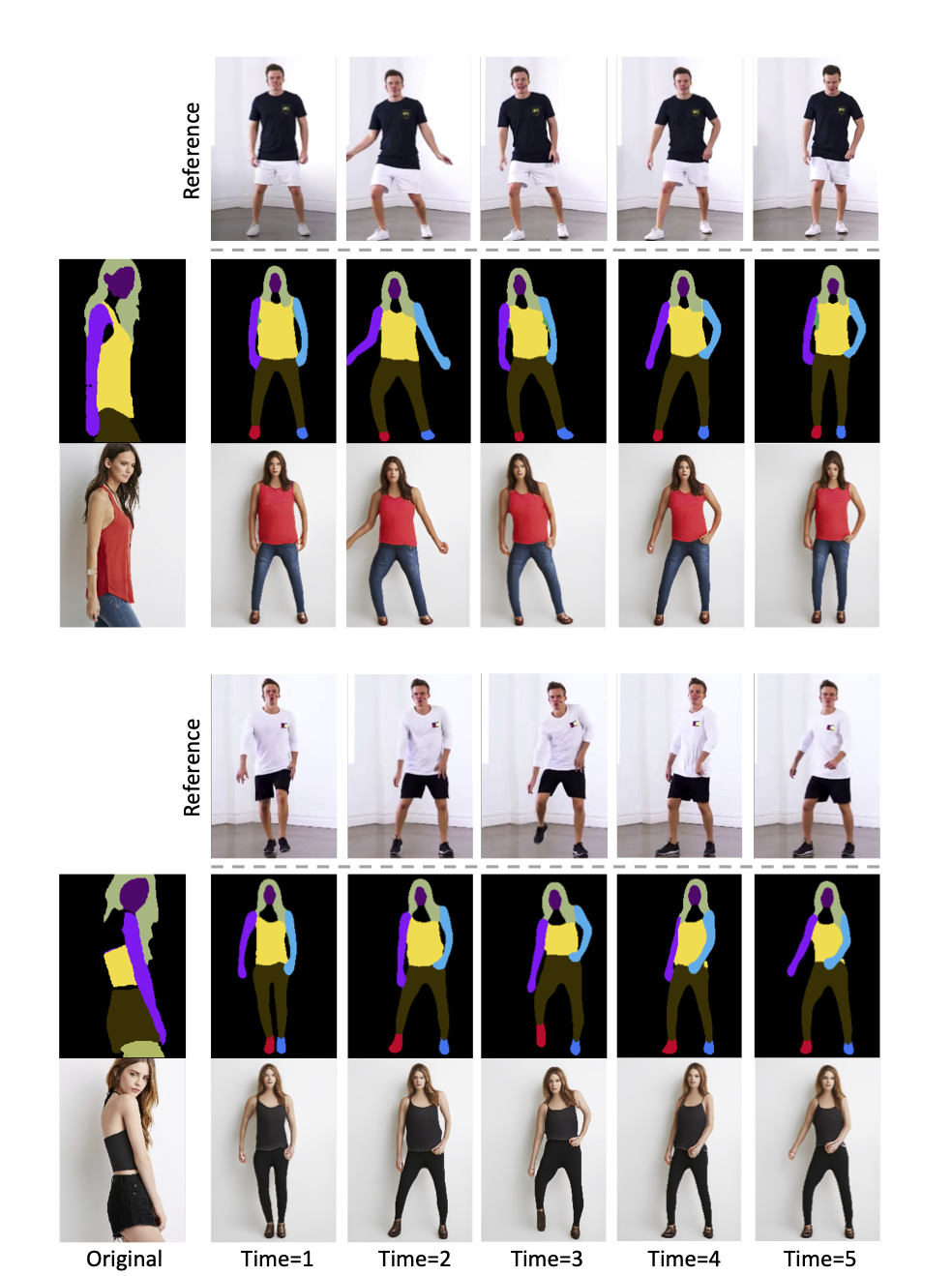}
  \end{center}
  \caption{Dancing as the reference with long pants. More video results are at \href{https://youtu.be/vTyq9HL6jgw}{\color{black}{https://youtu.be/vTyq9HL6jgw}}.} 
  \label{fig:video}
\end{figure*}

\end{document}